\documentclass[11pt,a4paper]{article}
\usepackage[hyperref]{conll-2019}
\usepackage{times}
\usepackage{latexsym}

\usepackage[utf8]{inputenc} 
\usepackage[T1]{fontenc}    
\usepackage{hyperref}       
\usepackage{url}            
\usepackage{booktabs}       
\usepackage{amsfonts}       
\usepackage{nicefrac}       
\usepackage{microtype}      
\usepackage{floatrow}
\usepackage{graphicx}
\usepackage{wrapfig}
\usepackage{placeins}
\usepackage{tablefootnote}
\usepackage{paralist}
\usepackage[font=small,labelfont=bf]{caption}

\usepackage{url}

\aclfinalcopy 

\title{Understanding Early Word Learning in Situated Artificial Agents}

\author{Felix Hill, Stephen Clark, Karl Moritz Hermann \& Phil Blunsom\\
  DeepMind \\
  London, UK \\
  \texttt{\{felixhill,clarkstephen,kmh,pblunsom\}@google.com} }

\date{}

\begin{document}
\maketitle
\begin{abstract}
Neural network-based systems can now learn to locate the referents of words and phrases in images, answer questions about visual scenes, and execute symbolic instructions as first-person actors in partially-observable worlds. To achieve this so-called grounded language learning, models must overcome challenges that infants face when learning their first words. While it is notable that models with no meaningful prior knowledge overcome these  obstacles, researchers currently lack a clear understanding of how they do so, a problem that we attempt to address in this paper. For maximum control and generality, we focus on a simple neural network-based language learning agent, trained via policy-gradient methods, which can interpret single-word instructions in a simulated 3D world. Whilst the goal is not to explicitly model infant word learning, we take inspiration from experimental paradigms in developmental psychology and apply some of these to the artificial agent, exploring the conditions under which established human biases and learning effects emerge. We further propose a novel method for visualising semantic representations in the agent.
\end{abstract}

\section{Introduction}

The learning challenge faced by children acquiring their first words has long fascinated philosophers and cognitive scientists \citep{Quine1960-QUIWO,brown1973first}. To start making sense of language, an infant must induce structure in a stream of continuous visual input, reconcile this structure with consistencies in the linguistic observations, store this knowledge in memory, and apply it to inform decisions about how best to respond to new utterances. 

Many neural network models also overcome a learning task that is -- to varying degrees -- analogous to early human word learning. Image classification tasks such as ImageNet \citep{deng2009imagenet} require models to induce discrete semantic classes, aligned to words, from unstructured pixel representations of large quantities of photographs \citep{krizhevsky2012imagenet}. Visual question answering (VQA) systems \citep{VQA,xiong2016dynamic,xu2016ask} must reconcile raw images with sequences of symbols, in the form of natural language questions, in order to predict lexical or phrasal answers. More recently, situated artificial agents have been developed that learn to understand sequences of words not only in terms of the contemporaneous raw visual input, but also in terms of past visual input and the actions required to execute an appropriate motor response \citep{oh2017zero,chaplot2017gated,hermann2017grounded,misra2017mapping}. The most advanced such agents learn to execute a range of phrasal instructions, such as~\textit{find the green object in the red room}, in a continous, simulated 3D world. To solve these tasks, an agent must execute sequences of hundreds of fine-grained actions, conditioned on the available sequence of language symbols and active (first-person) visual perception of the surroundings. 

The potential impact of situated linguistic agents is vast, as a basis for human users to interact with applications such as self-driving cars and domestic robots. However, our understanding of \textit{how} these agents learn and behave is limited. The challenges of interpreting the factors or reasoning behind the decisions and predictions of neural networks are well known. Indeed, a concerted body of research in both computer vision \citep{zeiler2014visualizing,Simonyan14a,yosinski2015understanding} and NLP \citep{linzen2016assessing, strobelt2016visual} is addressing this uncertainty. As grounded language learning agents become more prevalent, understanding their learning dynamics, representation and decision-making will become increasingly important, both to inform future research and to build confidence in users who interact with such models.  

\begin{figure*}[!t]
\hspace{5mm}
\noindent\makebox[\textwidth]{
  \includegraphics[width=14.5cm]{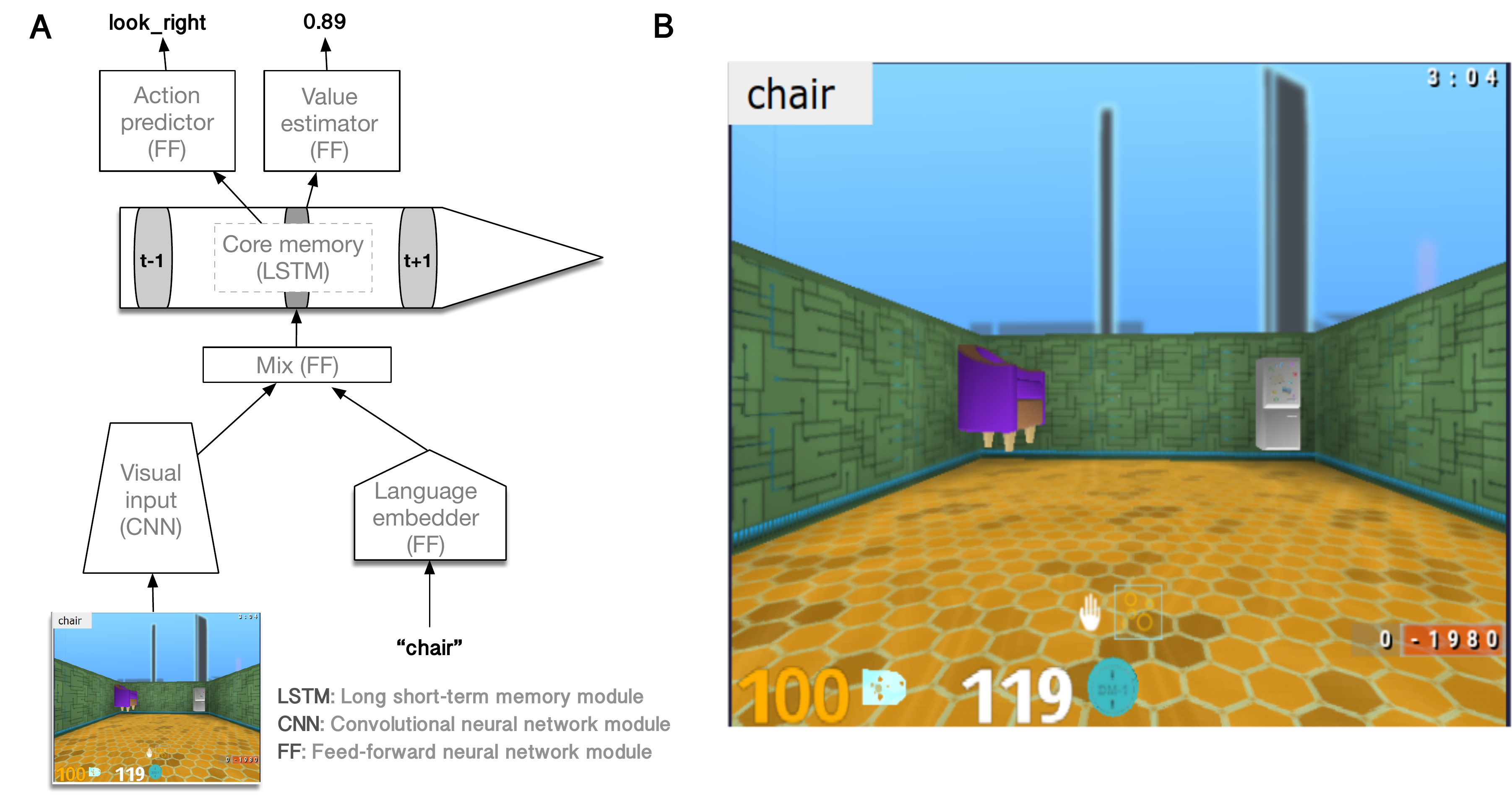}
}
\caption{\textbf{A:} Schematic agent architecture. \textbf{B:} An example of the word learning environment.}
\label{agent}
\vspace*{-0.1cm}
\end{figure*}

We aim to establish this understanding, noting the parallels with research in neuroscience and psychology that aims to understand human language acquisition. Extending the approach of~\newcite{ritter2017cognitive}, we adapt various techniques from experimental psychology \citep{landau1988importance,markman1990constraints,hollich2000breaking,colunga2005lexicon}. In line with typical human experiments, ours are conducted in a highly controlled environment: a simulated 3D world with a limited set of objects and properties, and symbolic linguistic stimuli (Fig.~\ref{agent}B). 
In each experimental episode, the agent is presented with a single word and two objects in a room. It must move by choosing between eight motor actions, viewing the objects from different perspectives until it can determine which one best reflects the meaning of the word.\footnote{There is more to knowing the meaning of a word than being able to identify an appropriate referent, but we are inspired by how infants initially learn to identify objects.} It receives a single scalar positive reward if it selects the correct object by moving towards and bumping into it. 

Even though our agent has a relatively generic architecture and uses a general-purpose learning algorithm, we show how it exhibits various aspects of early word learning. First, the agent successfully learns a  vocabulary of words from different semantic classes, and we study the dynamics of this process. We show that the rate at which the agent acquires new words increases rapidly after an initial slow period, an effect matching the human vocabulary spurt \cite{plunkett1992symbol,regier1996human}. We also propose two ways to speed up  word learning: moderating the agent's experience according to a curriculum \cite{elman1993learning} and an auxiliary learning objective reinforcing the association between words and the agent's replayed visual experience. 

Second, we investigate whether the agent exhibits a shape or colour bias \cite{brian1999emergence,regier2003emergent}. And finally, in order
to analyse the semantic processing taking place in the  model, we develop a novel method for dynamically visualising how different word types stimulate activations in different parts of the agent architecture. We also show that the agent clusters words according to the semantic classes of the word inputs \cite{elman1990finding,rumelhart1988parallel,hinton1981implementing}. 
   
\section{A 3D world for language learning}

\begin{table}[t]
\begin{tabular}{@{}lll@{}}
\toprule
\textbf{Word class} & \textbf{Example} & \textbf{Instruction meaning}\\ \textbf{(class size)}&& \textbf{(in this setting)}\\ 
\midrule
shapes (40)                                        & "\textit{pencil}" & Find and bump into \\ && a pencil.                      \\
colors (10)                                      & "\textit{blue}" & Find and bump into \\ &&  any blue object.                \\
patterns (2)                                        & "\textit{striped} "  & Find and bump into \\ &&  any striped object.              \\
relative                                  & "\textit{darker}"  & Find and bump into \\ shades (2) &&  the darker of the two \\ && objects in front of you.          \\
directions (2)                                     & "\textit{left}"     & Find and bump into \\ &&  the object furthest to \\ && the left as you look. \\
\bottomrule
\caption{Word classes learned by the situated agent.}
\label{tab:words}
\vspace*{-0.5cm}
\end{tabular}
\vspace*{-0.1cm}
\end{table}

Our experiments take place in the DeepMind Lab simulated world \citep{Beattie:2016}.\footnote{Open source version: https://github.com/deepmind/lab} During each episode, the agent receives a single word instruction (e.g. \textit{pencil\/}), and is rewarded for satisfying the instruction, in this case by executing  actions that allow it to locate a (3D, rotating) pencil and bump into it. At each time step in the episode, the agent receives a $ 3 \times 84 \times 84$ (RGB) pixel tensor of real-valued visual input, and a single word representing the instruction, and must execute a movement action from a set of 8 actions.\footnote{\texttt{\small move-forward, move-back, move-left, move-right, look-left, look-right, strafe-left, strafe-right}} The set of word classes, together with examples, is shown in Table~\ref{tab:words}. The total number of words is 56, chosen from 5 different classes. The full list is given in the Supplementary Material.

The episode ends after the agent bumps into any object, or when a limit of 100 timesteps is reached. To solve tasks and receive rewards, the agent must therefore first learn to perceive this environment, actively controlling what it sees via movement of its head (turning actions), and to navigate its surroundings via meaningful sequences of (typically around 60) fine-grained actions.

Figure~\ref{agent}B shows the experimental setup: the agent observes two 3D rotating objects and a language instruction word and must select the object that matches the instruction (in this case, a shape word (\textit{chair})). The confounding object (a refrigerator) and the colours of both objects are selected at random and vary across the agent's experience of the word \textit{chair} during training.

We fix the overall layout of the world (a rectangular room), the range of positions in which the agent begins an episode (towards the back of the room), the locations that objects can occupy (two positions at the front), a list of objects that can appear, the relative frequency of each object appearing and rewards associated with selecting a certain object given a particular instruction word. The environment engine is then responsible for randomly instantiating episodes that satisfy these constraints together with corresponding instruction words. Even with this relatively constrained level specification, there are an enormous number of unique episodes that the agent can encounter during training, each involving different object shapes, colours, patterns, shades, and/or relative positions. 

\section{A situated word-learning agent}

Our agent (Figure~\ref{agent}A), combines standard modules for processing symbolic input (an embedding layer) and visual input (a convolutional network).
At each time step $t$, the visual input $v_t$ is encoded by the
convolutional~\textit{vision module} and a \textit{language module} embeds the instruction word $l_t$.
A~\textit{mixing module} determines how these signals are combined
before they are passed to an LSTM~\textit{core memory}. In this work, the mixing module is simply a feedforward linear layer operating on the concatenation of the output from the vision and language modules, and the language module is a simple embedding lookup (since the instruction consists of one word). 

The hidden state $s_t$ of the core memory (LSTM) is fed to an action predictor (a fully-connected layer plus softmax), which computes the policy, a probability distribution over possible motor actions
$\pi(a_t|s_t)$, and a state-value function estimator $\mbox{\em Val\/}(s_t)$, which
computes a scalar estimate of the agent state-value function (the expected discounted future return). This value estimate is used to compute a baseline for the return in the asynchronous advantage actor-critic (A3C) policy-gradient algorithm \citep{mnih2016asynchronous}, which determines weight updates in the network in conjunction with the RMSProp optimiser \citep{tieleman2012lecture}. 

For an instantiation of an overall A3C agent, $16$ CPU cores each instantiate a single agent and a copy of the environment, applying gradient updates asynchronously to a centralised set of weights (which the agents share). Hyperparameters for the 16 instantiations are sampled from the ranges specified in the Supplementary Material, and we report the performance of the best 5 of these replicas. 

\section{Word learning dynamics}

\begin{figure*}[!t]
  \includegraphics[width=12.5cm]{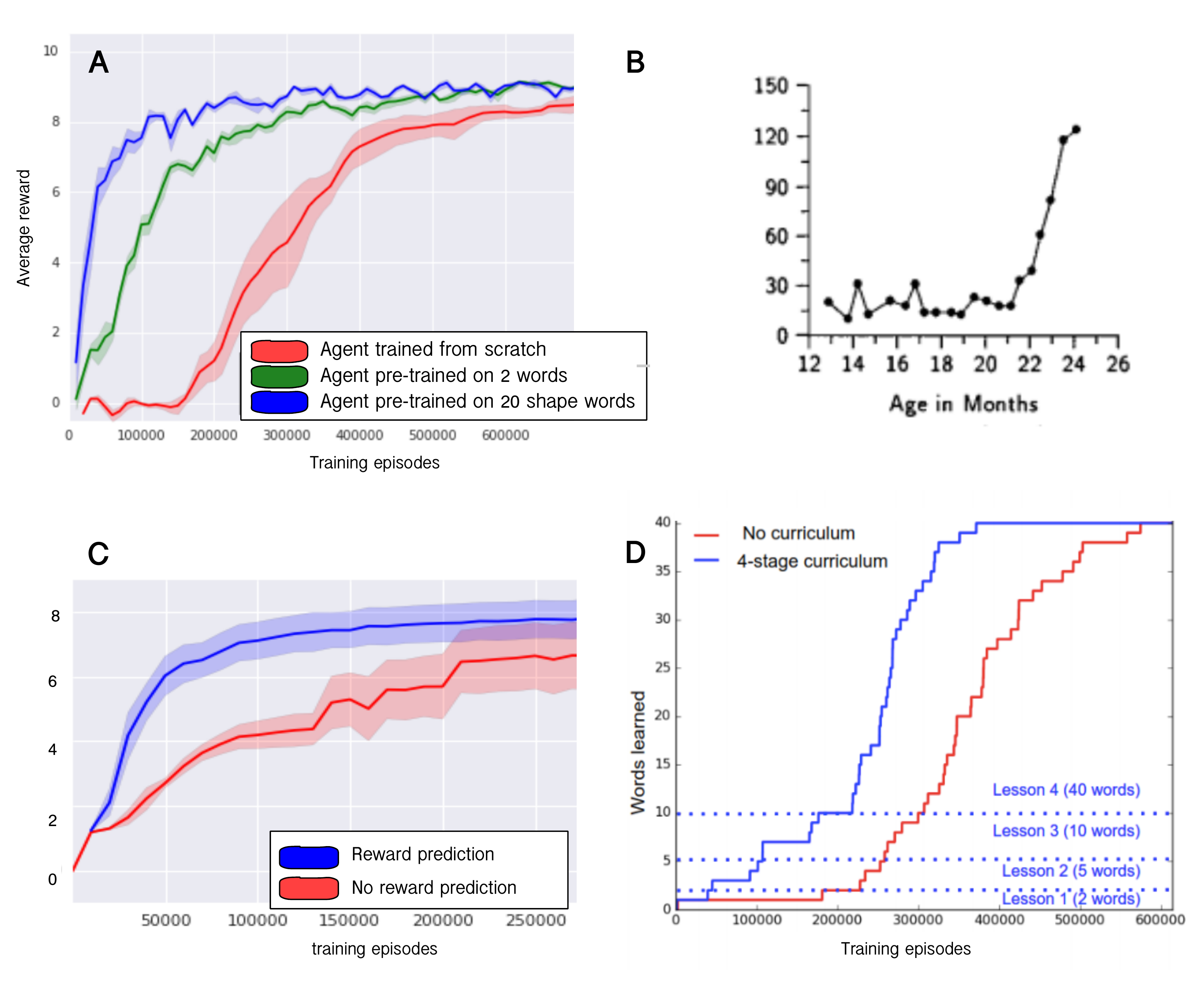}
\caption{\label{fig:speed}\textbf{Vocabulary growth in the agent:}~\textbf{A} Word learning trajectories for the A3C agent. \textbf{B} The sudden acceleration of vocabulary size in a human infant. \textbf{C} The effect of reward-prediction auxiliary loss on learning speed for an agent learning the full vocabulary of different word types.
\textbf{D} Word learning trajectories for an agent following a curriculum.}
\end{figure*}

In our first simulation, we randomly initialised all of the weights in the agent network, and then trained it on episodes with instruction words referring to the \textit{shape}, \textit{colour}, \textit{pattern}, \textit{relative shade} or \textit{position} of objects. Each episode began with the agent at one end of a small room and two objects at the other. Single instruction words were presented as discrete symbols at all timesteps. The instruction word in each episode unambiguously specified one of the two target objects, but other unimportant aspects of the environment could vary maximally. Thus, shape-word instructions could refer to objects of any colour, colour-words to objects of any shape, and so on. To eliminate ambiguity, when shade words \textit{dark}, \textit{light} were presented, both objects differed in shade but not colour, so that the meaning of the shade words was comparative (as in \textit{darker}, \textit{lighter}). The agent received a reward of $+10$ if it bumped into the correct object, $-10$ if it bumped into the wrong object, and $0$ if the maximum number of timesteps was reached. All words appeared with equal frequency during training.

We found that the agent slowly learned to respond correctly to the words it was presented with, but at some point the rate of word learning accelerated rapidly (Fig.~\ref{fig:speed}A, red curve). This effect is observed in both young infant learners \cite{nazzi2003before,dore1976transitional} and (supervised) connectionist simulations of word learning (Fig.~\ref{fig:speed}B, as recorded by~\newcite{plunkett1992symbol}). Our results show that the effect persists when such networks are trained with RL algorithms from raw pixel input. By the end of successful training, the agent was able to walk directly up to the two objects and reliably identify the appropriate referent.\footnote{For a video of an agent's behaviour, see \scriptsize{http://tiny.cc/m2nrty}.}

For our agents, some of the delay in the onset of word learning can be explained by the need to acquire relatively language-agnostic capacities such as useful sequences of motor actions or the distinction between objects and walls. However, some of the acceleration seems also to derive from the accruing semantic knowledge. To demonstrate this, we compared word learning speeds in an agent with prior knowledge of two words to an agent with knowledge of 20 words (Fig.~\ref{fig:speed}A, green and blue curves). The prior knowledge in this case was provided by training the agent on the word-learning task, as described above, but restricting the vocabulary to two and 20 words. So in both cases the agent has learned to "see" and move, but the agent pre-trained on 20 words learned new words more quickly. This effect accords with accounts of human development that emphasise how learning becomes easier the more the language learner knows \cite{bates1987competition}.  

We also explored ways to reduce the number of rewarded training episodes before word learning onset, in the form of a curriculum. We found one way to achieve this by moderating the scope of the learning challenge faced by the agent initially, before later expanding its experience once word learning had started. Specifically, we trained the agent to learn the meaning of the 40 shape words under two conditions. In one condition, the agent was presented with the 40 words (together with corresponding target and confounding objects) sampled randomly throughout training. In another condition, the agent was only presented with a subset of the 40 words (selected at random) until these were mastered (as indicated by an average reward of $9.8/10$ over 1000 consecutive trials), at which point this subset was expanded to include more words. So the stimuli are initially constrained to a two-word subset $S_1, S_1 \subset S$, until the agent learns both words, then extended to a 5-word subset $S_2, S_1 \subset S_2 \subset S$, then a 10-word subset $S_3,  S_2 \subset S_3 \subset S$, until finally being exposed to all 40 words in $S$.

As shown in Fig.~\ref{fig:speed}D, the agent following the curriculum reached 40 words faster than the agent confronted immediately with a large set of new words. This effect accords with the idea that early exposure to simple, clear linguistic input helps child language acquisition \citep{Fernald2010}. It also aligns with curriculum learning effects observed when training neural networks on text-based language data \cite{elman1993learning,bengio2009curriculum}.  

We found a further way to reduce the number of episodes required to achieve word learning by applying an auxiliary learning objective on stored trajectories of the agent's experience, in a manner proposed by \newcite{jaderberg2016reinforcement} (Fig.~\ref{fig:speed}C).\footnote{Data in this and other learning curves show the best 5 += SE from 16 replicas launched with hyperparameters sampled from ranges specified in the Supp. Material.} In agents with this auxiliary prediction process, the final 4 observations of each episode are saved in a replay buffer and processed offline by the visual and language modules. The concatenation of the output of these modules is then used to predict whether the episode reward was positive, negative or zero. A cross-entropy loss on this prediction is optimised jointly with the agent’s A3C loss. 

This application of an auxiliary prediction loss can be seen as a rudimentary model of hippocampal replay biased towards rewarding events, a mechanism that is thought to play an important role in both human and animal learning \cite{schacter2012future,gluck1993hippocampal,pfeiffer2017content}. The auxiliary loss serves to reinforce the correspondence between visual scenes and words by effectively posing the question \textit{does this word match this view?}. This internal question-answering process seems to complement the instruction following, leading to  faster word learning at early stages.

\section{Word learning biases}

\begin{figure*}[!t]
\noindent\makebox[\textwidth]{\includegraphics[height=5cm]{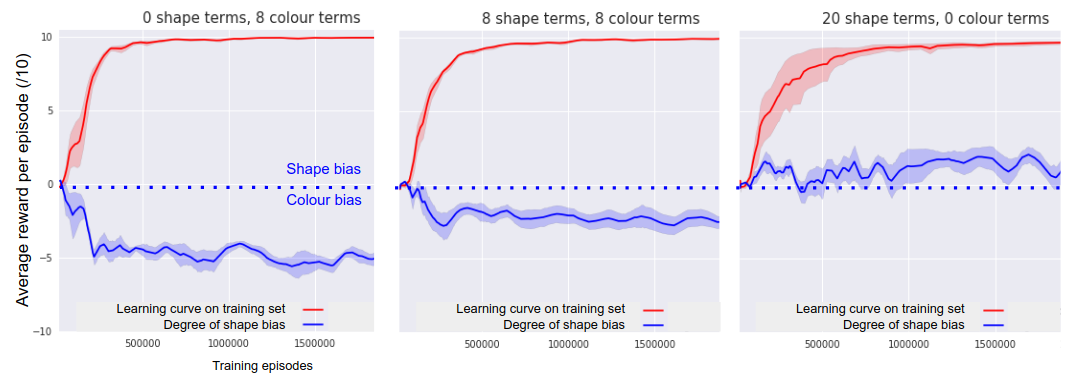}}
\caption{\label{fig:shape_bias} Degrees of shape bias for different training regimes.}
\end{figure*}

It is widely agreed that children exploit certain labelling biases during early word learning, which serve to constrain the possible referents of novel, ambiguous lexical stimuli \cite{markman1990constraints}. \newcite{regier2003emergent} discusses in detail various accounts of how such constraints or biases can emerge naturally from environment signals in connectionist models. A particularly well-studied learning constraint is the \textit{shape bias} \citep{landau1988importance}, whereby infants tend to presume that novel words refer to the shape of an unfamiliar object rather than, for instance, its colour, size or texture. Our simulated environment permits faithful replication of the original experiments by \newcite{landau1988importance} that uncovered the shape bias in infants. 

During training, the agent learns word meanings in a room containing two objects, one that matches the instruction word (positive reward) and a confounding object that does not (negative reward). Using this method, the agent is taught the meaning of a set $C$ of colour terms, $S$ of shape terms and $A$ of ambiguous terms (in the original experiment, the terms $a \in A$ were the nonsense terms `dax' and `riff'). The target referent for a shape term $s \in S$ can be of any colour $c \in C$ and, similarly, the target referent when learning the colours in $C$ can be of any shape. In contrast, the ambiguous terms in $A$ always correspond to objects with a specific colour $c_a \notin C $ and shape $s_a \notin S$ (e.g. `dax' always refers to a black pencil during training, and neither black nor pencils are observed in any other context).\footnote{Here colour terms refer to a range of RGB values through application of Gaussian noise to prototypical RGB codes, so two instances of red objects will have subtly different colours.} 

As the agent learns, we periodically measure its bias by means of test episodes for which no learning takes place. In a test episode, the agent receives an instruction $a \in A$ (e.g. `dax') and must decide between two objects, $o_1$, whose shape is $s_a$ and whose colour is $\hat{c} \notin C \cup \{c_a\}$ (e.g. a blue pencil), and $o_2$, whose shape is $\hat{s} \notin S \cup \{s_a\}$ and whose colour is $c_a$ (e.g. a black fork). Note that in the example neither the colour~\textit{blue} nor the shape~\textit{fork} are observed by the agent during training. As with the original human experiment, the degree of shape bias in the agent can be measured, as the agent is learning, by its propensity to select $o_1$ in preference to $o_2$. Moreover, by varying the size of sets $S$ and $C$, we can examine how different training regimes affect the bias exhibited by the agent.

Fig.~\ref{fig:shape_bias} illustrates how a shape/colour bias develops in agents exposed to three different training regimes. The bias is represented by the blue line, which is the mean "score" when +10 is awarded for the object matching the instruction in shape, and -10 for the object matching in colour, over 1000 random test episodes (i.e. a line below zero indicates a propensity to choose objects matching in colour). An agent that is taught exclusively colour words ($| S | = 0$, $| C | = 8$) unsurprisingly develops a strong colour bias. More interestingly, an agent that is taught an equal number of shape and colour terms ($| S | = 8$, $| C |= 8$) also develops a colour bias. In order to induce a (human-like) shape bias, it was necessary to train the agent exclusively on a larger set of ($| S | = 20$, $| C | = 0$) shapes before it began to exhibit a notable shape bias.  

It is notable that in the balanced condition our agent architecture (convolutional vision network combined with language instruction embedding) naturally promotes a colour bias. This may be simply because, unlike information pertinent to shapes, the agent has direct access to colour in the RGB stream of pixel input, so that if the environment is balanced, specialising perceptual and grounding mechanisms in favour of colours is a more immediate path to higher returns. Note also that our conclusion differs from that of \newcite{ritter2017cognitive}, who observed a shape bias in convolutional networks trained on ImageNet. Our experiments suggest that this effect is more likely driven by the distribution of training data (the ImageNet data  contains many more shape-based than colour-based categories) rather than the underlying convolutional architecture. Indeed, in the present model, it may be that this flexible ability to induce relevant biases facilitates the sudden acceleration of word learning described earlier. As the agent's object recognition and labelling mechanisms specialise (towards shapes, colours or both, as determined by the environment), the space of plausible referents for new words narrows, permitting faster word learning as training progresses.

Regarding human learners, our simulations accord with accounts of the shape bias that emphasise the role of environmental factors in stimulating the development of such a bias \cite{regier2003emergent}. The fact that shape terms occur with greater frequency in typical linguistic environments, for American children at least, can be verified by analysis of the child-directed language corpus Wordbank \citep{frank2017wordbank}. Our findings indicate that the prevalence of the human shape bias could be as much a product of the prevalence and functional importance of shape categories in the experience of typical infants as a reflection of the default state of their underlying perceptual and cognitive mechanisms.

\section{Visualising grounding in both action and perception}
\label{sec:structure}

One compelling aspect of early word learning in humans is infants' ability to make sense of apparently unstructured raw perceptual stimuli. This process requires the learner to induce meaningful extensions for words (when there are limitless potential referents in the environment), and to organise these word meanings in semantic memory. The success of this process has been explained by innate cognitive machinery delimiting conceptual domains, or at least for narrowing the space of possible referents \cite{marcus1999poverty}. Alternative accounts, which accord more closely with the learning mechanism presented here, emphasise the capacity of associative learning systems to infer word meanings by exploiting diverse signals in the environment, and bootstrapping currently known words to learn new words more easily \cite{elman1990finding,smith2008infants,Frank2013}.

\begin{figure}[]
  \centering
    \includegraphics[width=0.95\textwidth,trim={5em 0 9em 0},clip]{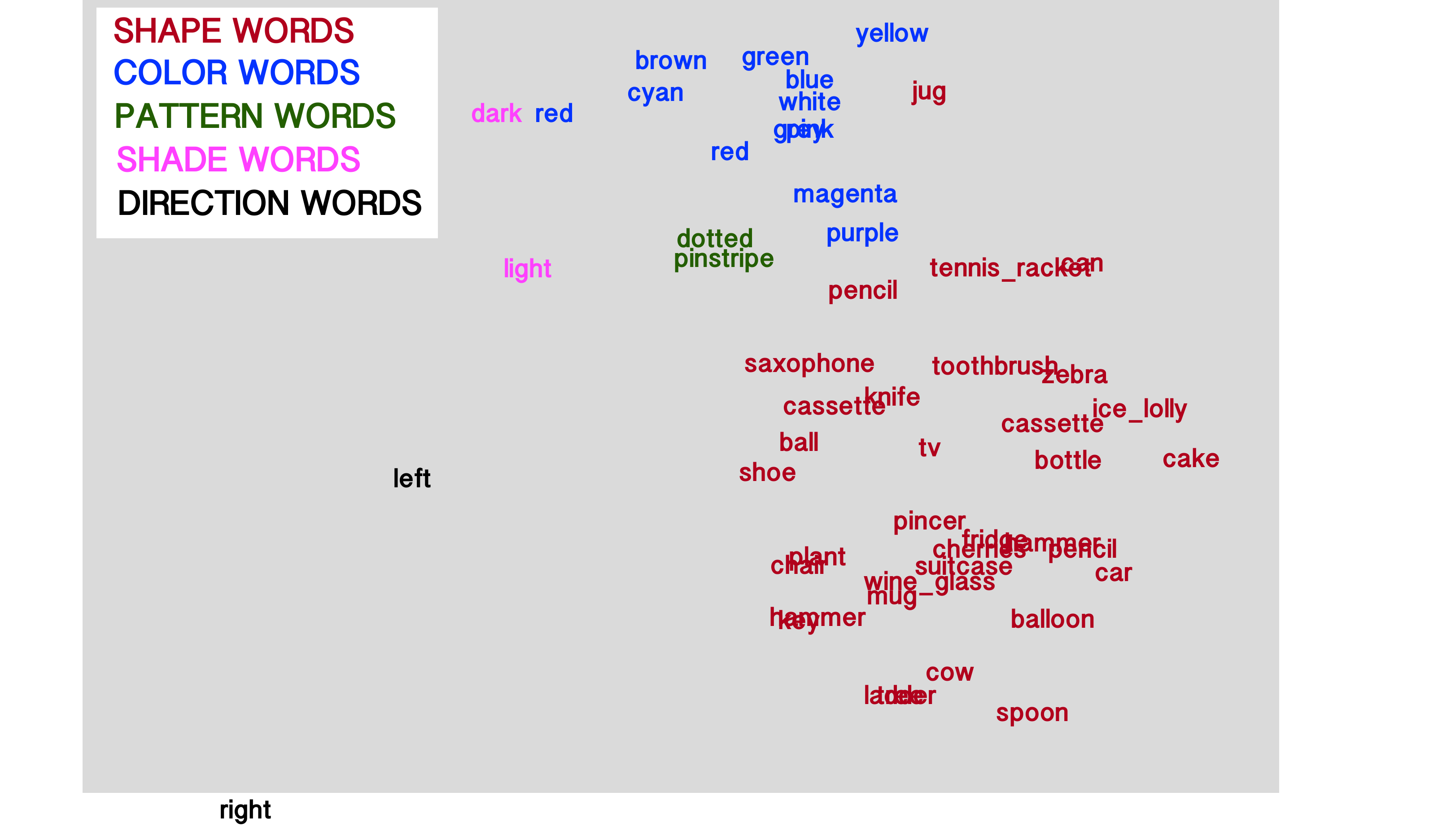}
    \caption{\label{fig:embeddings} t-SNE projection of semantic and syntactic (adjective/noun) classes in the agent's word representation space.}
    \vspace*{-0.2cm}
\end{figure}

\begin{figure*}[!t]
\noindent\makebox[\textwidth]{\includegraphics[width=0.85\linewidth]{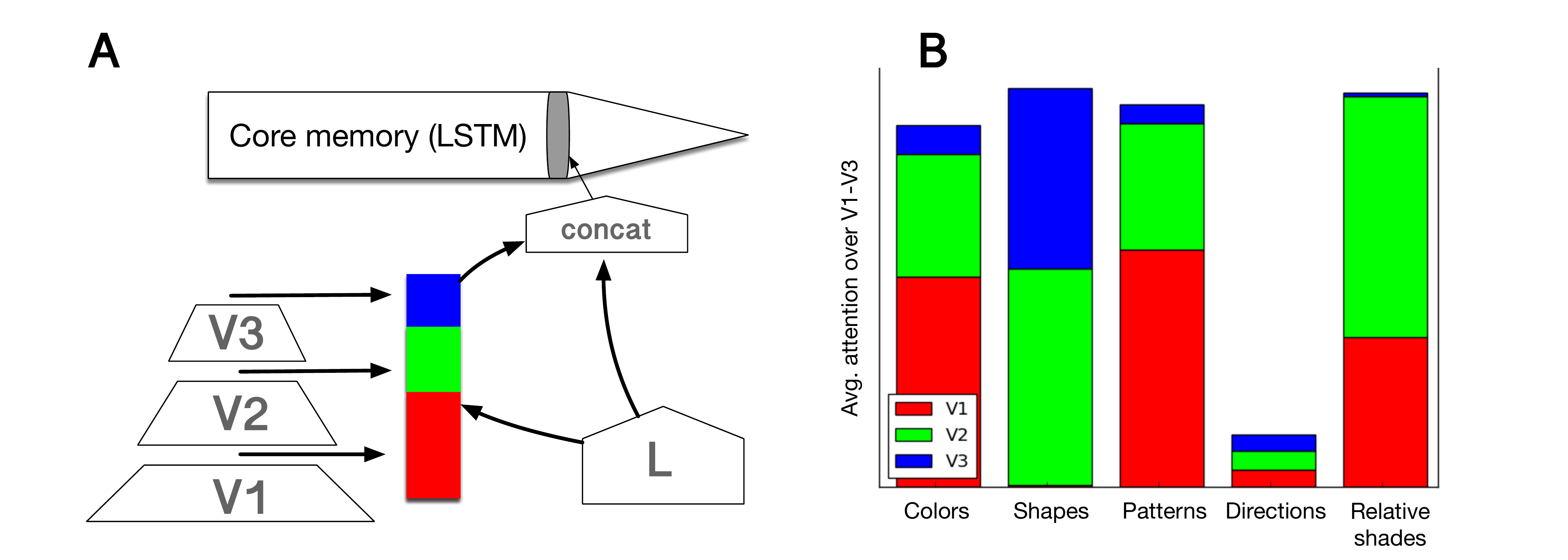}}
\caption{\label{fig:attention}\textbf{Online semantic processing in the agent.} \textbf{A}: The modified agent network with a layerwise attention module. \textbf{B}: Average activations flowing through modified agents trained on words of different types, averaged over 100 episodes.}
\vspace*{-0.2cm}
\end{figure*}

We analysed the trained agent to better understand how it solves the problem of cross-situational word learning in our setting. First, we visualised the space of word embeddings in an agent trained on words from the different classes detailed in Fig. \ref{fig:embeddings}, with experience sampled uniformly over words, not classes. We observe that these word classes, which align with both semantic (shape vs. color) and syntactic (adjective vs. noun) categories, emerge naturally in the embedding space of the agent as it discovered the underlying relationship between words, raw-pixel visual observations of the environment and the `correct' set of referents as encoded in the environment design.

We further explored how this emergent semantic structure manifested itself in processing across the network during an episode. To do so, we adapted the network to compute weightings for visual field locations at all layers of its visual processing module (a modification we term \textit{layerwise attention}), and measured these weights when agents were trained to understand words of the different types. 

More precisely, let $e_l$ be the representation of an instruction word $l$ and $\mathbf{v}_i$ be the output of layer $i = 1, 2, 3$ of the visual module with dimension $ n_i \times n_i \times k_i $, where $k_i$ is the number of feature maps. In the layerwise attention module, the $\mathbf{v}_i$ are first passed through 3 independent linear layers to $\mathbf{v'}_i$ with common final dimension $ n_i \times n_i \times K $, such that $K$ is also the dimensionality of $e_l$. The $\mathbf{v'}_i$ are then stacked into a single tensor $T$ of dimension $ d \times K $, where $d = \sum_{i=1}^3 n^2_i$. $T$ is then multiplied by $e_l$ and passed through a softmax layer to yield a $d$ dimensional discrete probability distribution over all (pixel-like) locations represented in each layer of the visual module $\mathbf{V}$. These values are applied in a weighted sum of the ($k_i$-dimension) representations returned by each layer before concatenation, as before, with $e_l$.  

By analysing the distribution over spatial locations and visual layers computed by the layerwise attention mechanism, we found that colour and shade words words stimulated activations at the lower levels of the visual-processing module, whereas shape word stimuli activated comparatively more features computed at higher levels (see the red, green and blue bars in Fig~\ref{fig:attention}B, showing activations at levels 1, 2 and 3 of the CNN, respectively). This observation accords with previous analyses of filters in convolutional networks for trained image classification \cite{lecun2010convolutional}. 

At the mixing (concatenation) layer of the network, we also measured the relative strength of total activation flowing through the visual versus linguistic pathways for agents trained on different word types, and observed that the direction words were associated with much lower activations from the visual module than other word types. (See the total height of the bars in Fig~\ref{fig:attention}B, which indicates relative activation strength on visual vs. language units, so the higher the bar the more avitvation on the visual side.) This observation underlines the embodied nature of representation in the agent. Effectively, direction words are grounded in actions to a greater extent than vision, a finding that aligns with cognitive and neuroscientific theories that emphasise the interaction between linguistic semantic representation and sensory-motor processes \cite{pulvermuller2014motor}.

\begin{figure}[ht]
\vspace*{0cm}
  \begin{center}
    \includegraphics[width=\textwidth]{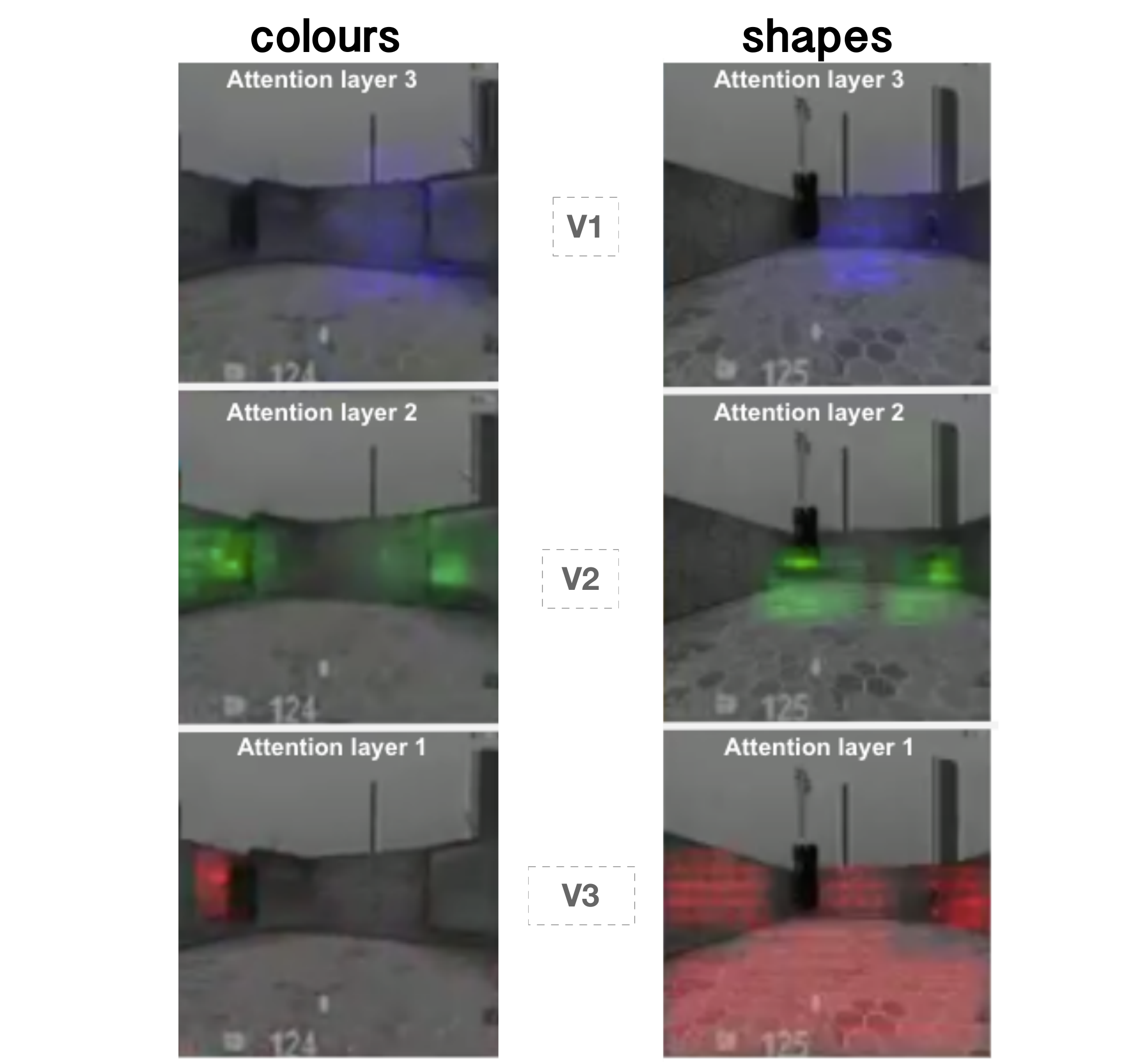}
    \caption{\label{attn}The attention weighting of different spatial locations V1-V3, backpropagated to the visual field.}
      \end{center}
      \vspace*{-0.2cm}
\end{figure}

Finally, these quantitative analyses of the activations can be backpropagated onto the agent visual input, as per \newcite{Simonyan14a}, to visualise the focus of this attention, as illustrated in Fig~\ref{attn}.\footnote{ \protect\texttt{\scriptsize{http://tiny.cc/m2nrty}} is a video of the agent attention.}

\section{Discussion and conclusions}

Recent work has demonstrated the power of training deep RL agents conditioned on language input \cite{hermann2017grounded, chaplot2017gated,oh2017zero,bahdanau2018learning,misra2018mapping,fu2019language,das2018neural}. Here we have studied an end-to-end model of cross-situational word learning that can ground word meanings in (active-)perception and actions, while relying on few prior assumptions about representation of the visual environment or cognitive states.

We have explicitly stated that the theme of this paper is technological: our aim is to develop embodied agents with linguistic capabilities in simulated environments, with the ultimate goal of transferring such agents to the real world. Clearly the linguistic capacities of this particular agent are rudimentary, but we have focused on early word learning in order to apply methods from developmental psychology as a means to analyse the agent's representations and behaviour. Note also that we have chosen to focus on a 3D environment, rather than, say, images, because of the potential of such environments to develop useful technology in the future. It is possible that our techniques could be applied to the VQA setting, also.

Whilst we are not offering the agent here as an accurate model of human word learning, we have on occasion suggested how our findings might accord with, or support, theories from the language developmental literature. And we do suggest that taking such embodied agents seriously as models of human language learning could be productive from a scientific point of view. Compared with prior models of early word learning, our situated agent is novel in its ability to learn jointly, end-to-end from word stimuli and raw pixel perception to behaviour. It is also possible that, by considering how such agents are currently deficient when it comes to modeling human learning, we could improve the technology. Hence here are some ways in which our current setup is unlike that faced by the infant language learner.

Since our model receives symbolic lexical stimuli, it does not need to segment the speech stream \cite{roy2002learning}, or isolate words from natural (multi-word) child-directed speech \cite{larsen2017relating}. Further, unlike young children, the model does not learn a new label for an object after a single experience (so-called fast-mapping) \cite{xu2007word}. The existence of only a single agent in our present environment makes it impossible for the models to apply pragmatic inference or exploit social cues \cite{Frank2013}, and the visual complexity does not match that of the real world \cite{ritter2017cognitive}. Finally, the predominant learning setting for our agent is (rewarded) instruction following, which is a frequent experience for language learners in certain cultures, but rare in others \cite{cristia2017child}.

In future work we therefore hope to develop end-to-end models that can jointly apply multiple different mechanisms, exploiting cues from even richer sets of input streams, in order to deal with more of the phenomena that are commonly observed when infants learn their first words, and which will be needed in future technological applications. 

\section*{Acknowledgements}
We would like to thank Michael C Frank, Noah Goodman, Jay McClelland, Sam Ritter, Hinrich
Sch\"utze and all attendees of the MIC3 workshop for many helpful ideas and suggestions.

\bibliography{conll-2019}
\bibliographystyle{acl_natbib}

\end{document}


\section*{Supplementary Material}
\label{App}

\section{Agent details}
\subsection{Fixed agent hyperparameters} 
See Table~\ref{tab:exp1}.
\begin{table*}[ht]
\scriptsize
\centering
\begin{tabular}{lll}

\textbf{Hyperparameter}              & \textbf{Value}       & \textbf{Description}   \\ 
train\_steps                & 640m & Theoretical maximum number of time steps (across all episodes) for which the agent will be trained. \\
env\_steps\_per\_core\_step & 4           & Number of time steps between each action decision (action smoothing)\\
num\_workers & 32   & Number of independent workers running replicas of the environment with asynchronous updating.  \\
unroll\_length   & 50  & Number of time steps through which error is backpropagated in the core LSTM action module   \\
& &\\
\bf{visual encoder}  &   &  \\
num\_layers               & 3         & Layers in the convolutional vision network.                                                      \\
output\_channels              & (32, 64, 64)        & Number of feature maps in each layer of the network.                                                      \\
kernel\_shapes             & (8, 4, 3)        & Shapes of the (square) convolutional kernels in each layer of the network.                                                      \\
strides              & (4, 2, 1)       & Convolution stride length in each layer of the network.                                                      \\
activation              & relu      & Activation function applied after all except the final layer of the visual encoder.                                                      \\
                            &             &         \\
\bf{language encoder}                &             &                                                                                                                                                \\
embedding\_dim              & 128        & Dimension of the word and instruction embeddings. \\
                            &             &         \\
\bf{cost calculation}                        &             &                                                                                                                                                \\
additional\_discounting     & 0.99        & Discount used to compute the long-term return R\_t in the A3C objective                                                                        \\
cost\_base                  & 0.5         & Multiplicative scaling of all computed gradients on the backward pass in the network                                                           \\
                            &             &                                                                                                                                                \\
\bf{optimisation}             &             &                                                                                                                                                \\
clip\_grad\_norm            & 100         & Limit on the norm of the gradient across all agent network parameters (if above, scale down)                                                   \\
decay                       & 0.99        & Decay term in RMSprop gradient averaging function                                                                                              \\
epsilon                     & 0.1         & Epsilon term in RMSprop gradient averaging function                                                                                            \\
learning\_rate\_finish      & 0           & Learning rate at the end of training, based on which linear annealing of is applied\\
momentum    & 0         & Momentum parameter in RMSprop gradient averaging function  \\ 
\end{tabular}
\caption{\label{tab:exp1}Agent hyperparameters that are fixed throughout our experimentation.}
\end{table*}

\subsection{Variable agent hyperparameters}

Agent hyperparameters randomly sampled in order to yield different replicas of our agents for training. \textit{uniform($x, y$)} indicates that values are sampled uniformly from the range $[x, y]$. \textit{loguniform($x, y$)} indicates that values are sampled from a uniform distribution in log-space (favouring lower values) on the range $[x, y]$. Unless otherwise stated, 16 replicas were started for each experimental condition, and results reported reflect the best 5 of these replicas.

See Table~\ref{tab:exp2}.

\begin{table*}[h]
\scriptsize
\centering
\begin{tabular}{lll}
\textbf{Hyperparameter}       & \textbf{Value}                    & \textbf{Description} \\ 
\bf{language encoder} &   & \\
embed\_init           & \textit{uniform(0.5, 1)}           & Standard deviation of normal distribution (mean = 0) for sampling \\
&& initial values of word-embedding weights in\bf{ L}\\
&                           &\\
\bf{optimisation}          &                           &\\
entropy\_cost         & \textit{uniform(0.0005, 0.005)}    & Strength of the (additive) entropy regularisation term in the A3C cost function.                                               \\
learning\_rate\_start & \textit{loguniform(0.0001, 0.002)} & Learning rate at the beginning of training \\
&& annealed linearly to reach learning\_rate\_finish at the end of train\_steps. \\
\end{tabular}
\label{tab:hypB}
\caption{\label{tab:exp2}Agent hyperparameters that randomly sampled in order to yield    different replicas of our agents for training. \textit{uniform($x, y$)}    indicates that values are sampled uniformly from the range $[x, y]$.    \textit{loguniform($x, y$)} indicates that values are sampled from a    uniform distribution in log-space (favouring lower values) on the    range $[x, y]$.  }
\end{table*}

\section{Experiment details}
The full set of words used. See Table~\ref{tab:words}.

\begin{table*}[!h]
    \footnotesize
    \begin{tabular}{ll}
   \textbf{ Word class} &  \textbf{Words} \\ 
    Shapes (40) & \textit{chair, suitcase, tv, ball, balloon, cow, zebra} 
     \textit{cake, can, cassette, chair, guitar, hair-brush,}\\
     & \textit{hat, ice-lolly, ladder, mug, pencil, suitcase,} 
     \textit{toothbrush, key, bottle, car, cherries, fork,} \\
     & \textit{fridge, hammer, knife, spoon, apple, banana} 
     \textit{flower, jug, pig, pincer, plant, saxophone,}  \\
     & \textit{shoe, tennis-racket, tomato, tree, wine-glass} \\
    Colours (10) & \textit{blue, brown, pink, yellow, red, green, cyan, magenta, grey, purple} \\
    Patterns (2) & \textit{pinstriped, dotted} \\
    Shades (2) & \textit{lighter, darker} \\
    Directions (2) & \textit{left, right} \\
    \end{tabular}
    \caption{\label{tab:words}Full list of words.}
\end{table*}